\documentclass{article}

\usepackage{microtype}
\usepackage{graphicx}
\usepackage{subfigure}
\usepackage{booktabs} 
\usepackage{graphicx}
\usepackage{hyperref}
\usepackage{multirow}

\usepackage[accepted]{icml2020}
\usepackage{extra_style}

\usepackage{xcolor}

\icmltitlerunning{Neural Ordinary Differential Equations on Manifolds}

\begin{document}

\twocolumn[
\icmltitle{Neural Ordinary Differential Equations on Manifolds}



\icmlsetsymbol{equal}{*}

\begin{icmlauthorlist}
\icmlauthor{Luca Falorsi}{to,goo}
\icmlauthor{Patrick Forr\'e}{to}
\end{icmlauthorlist}

\icmlaffiliation{to}{University of Amsterdam}
\icmlaffiliation{goo}{Aiconic}

\icmlcorrespondingauthor{Luca Falorsi}{luca.falorsi@gmail.com}

\icmlkeywords{TODO, ADD, KEYWORDS}

\vskip 0.3in
]



\printAffiliationsAndNotice{} 

\begin{abstract}
Normalizing flows are a powerful technique for obtaining reparameterizable samples from complex multimodal distributions. Unfortunately current approaches fall short when the underlying space has a non trivial topology, and are only available for the most basic geometries. Recently normalizing flows in Euclidean space based on Neural ODEs show great promise, yet suffer the same limitations. 
Using ideas from differential geometry and geometric control theory, we describe how neural ODEs can be extended to smooth manifolds. We show how vector fields provide a general framework for parameterizing a flexible class of invertible mapping on these spaces and we illustrate how gradient based learning can be performed. As a result we define a general methodology for building normalizing flows on manifolds. 
\end{abstract}

\section{Introduction}
Recently \citet{NIPS2018_7892} showed how to effectively integrate Ordinary Differential Equations (ODE) with Deep Learning frameworks. 
Ubiquitous in all fields of science, differential equations are the main modelling tools for physical processes. In deep learning their introduction was initially motivated from the observation that the popular residual network (ResNet) architecture can be interpreted as an Euler discretization step of a differential equation \citep{haber2017stable}. Instead of relying on discretized maps,
\citet{NIPS2018_7892} proposed to directly model the continuous dynamics using \emph{ vector fields} in $\bbR^n$. A vector field, through its associated ODE, indicates for every point an infinitesimal displacement change, and therefore implicitly describes a map from the space to itself called a \emph{flow}. While the flow can be practically computed using numerical ODE solvers, the key observation of \citet{NIPS2018_7892} is that we can threat the ODE solution as a black-box. This means that in the backward pass we do not have to differentiate through the operations performed by the numerical solver, instead \citet{NIPS2018_7892} propose to use the adjoint sensitivity method \citep{pontryagin1962mathematical}. 
Closely related with the Pontryagin Maximum Principle, one of the most prominent results in control theory, the adjoint sensitivity method allows to compute vector-Jacobian product of the ODE solutions with respect of its inputs. This is done by simulating the dynamics given by the initial ODE backwards, augmenting it with a linear differential equation, which intuitively can be thought of as a continuous version of the usual chain rule. 

Neural ODEs found one of their major applications in the context of {\it normalizing flows} \citep{grathwohl2018scalable, finlay2020train}. Normalizing flows are a general methodology for defining complex reparameterizable densities by applying a series of diffeomorphism to samples from a (simple) base distribution \citep{rezende2015variational}\footnote{See \citep{papamakarios2019normalizing} for a general review of NF}. The resulting density at the transformed points can be computed using the change of variable formula. Flows defined by vector fields are particularly amenable for this task, as for every time interval the ode solution defines a diffeomorphism. In this case, the change change of density is given simply by integrating the divergence of the vector field along integral curves. 

As many real word problems are naturally defined on spaces with a non-trivial topology, recently there has been a great interest in building probabilistic deep learning frameworks that can works on manifolds different from the Euclidean space \citep{davidson2018hyperspherical, falorsi2018explorations, falorsi2019reparameterizing, diffusionvae, wrappedhyperbolic}. 
For this objective the possibility of defining complex reparameterizable densities on manifolds through normalizing flows is of central importance. However as of today there exist few alternatives, mostly limited to the most basic and simple topologies. 

The main obstacle for defining normalizing flows on manifolds is that there is no general methodology for parameterizing maps $F:M\to N$ between two manifolds. Neural networks can only accomplish this for the the Euclidean space, $\bbR^n$. In this work we propose to use vector fields on a manifold $M$ as a flexible way to parameterize diffeomorphic maps from the manifold to itself. As a vector fields defines an infinitesimal displacement on the manifold for every point, they naturally give rise to diffeomorphisms without needing to impose further constraints. In addition, there exist decades old research on how to numerically integrate ODEs on manifolds\footnote{See \citet{hairer2011solving} for a review of the main methods}. 

We start in Section \ref{sec:vec-flow} by delineating how vector fields and ODEs on a manifold $M$ can be defined in the context of differential geometry. We then explain how vector fields naturally give rise, through their associated flow, to diffeomorphisms on $M$. In Section \ref{sec:cot-lift} we describe how the adjoint sensitivity method can be generalized to vector fields on manifolds in the context of geometric control theory \cite{agrachev2013control}. This highlights important connections with symplectic geometry and the Hamiltonian formalism. Similarly as in the adjoint method in the Euclidean space, to backpropagate through the flow defined by a vector field we have to solve an ODE in an augmented space. In this case the ODE is given by a vector field on the cotangent space $T^*M$, called \emph{cotangent lift}, which is a {\it lift} of the original vector field on $M$. In Section \ref{sec:cnf} we demonstrate how flows defined by vector fields allow to define \emph{continuous normalizing flows} on manifold. As a proof of concept we show how the defined framework can be used to model complex multimodal densities on the hypersphere in Appendix \ref{app:experimments} and provide practical advice on how to parameterize vector fields on a manifold using neural networks in Appendix \ref{app:par-field}. 

\section{Vector fields and flows on manifolds}\label{sec:vec-flow}

Throughout the paper $M$ will be a $n$-dimensional smooth manifold. Vector fields are the mathematical object that allows us to generalize the concept of ODEs to manifolds. A {\bf smooth vector field} $X$ is defined as a smooth section of the tangent bundle $X\in\smoothsec{M}{TM}$. A smooth time dependent vector field is a smooth function $X: \bbR\times M \to TM$ such that $\forall t\in \bbR$, $X_t:=X(t, \cdot)\in \smoothsec{M}{TM}$ is a smooth vector field. 
\begin{defn}
Let $X:\bbR\times M \to TM$ be a smooth time dependent vector field and $J\subseteq \bbR$ an interval. A curve $\gamma:J\to M$ is an {\bf integral curve} of $X$ if:
\begin{equation}\label{eq:integral-curve}
    \dot{\gamma}(t)=X(t,{\gamma(t)})\quad \forall t\in J
\end{equation}
We call {\bf maximal integral curve} an integral curve that cannot be extended to a larger interval $J$.
\end{defn}
Writing Equation \eqref{eq:integral-curve} in a local smooth chart we find that it is equivalent to (locally) solving a system of ODEs. We can then apply the existence and uniqueness theorem to show that every smooth vector field always admits integral curves:

\begin{thm}[Theorem 2.15 in \citep{AgrBarBos17}]
Let $X$ a smooth time dependent vector field, then for any point $(t_0,p_0)\in \bbR\times M$ there exist a unique maximal integral curve $\gamma:t_0\in J\to M$ with starting point $q_0$, and starting time $t_0$ denoted by $\gamma(t;t_0,q_0)$. We call $\gamma$ a solution of the Cauchy problem:
\begin{align}\label{eq:cauchy-problem}
    \left\{
    \begin{array}{l}
    \dot q(t) = X(q(t)) \\
    q(t_0) = q_0
    \end{array}
    \right.
\end{align}
Moreover the map $(t_0, q_0) \to \gamma(t; t_0, q_0)$ is smooth on a neighborhood of $(t_0, q_0)$. 
\end{thm}
A time-dependent vector field is {\bf complete} if for every $(t_0, q_0)\in \bbR\times M$, the maximal solution $\gamma(t;t_0,q_0)$ of the Cauchy problem is defined on all $\bbR$. 

Through integral curves, vector fields on manifolds give us a flexible and convenient way of defining maps from $M$ to itself. 
Restricting to the time independent case, this means considering the family of maps $\phi_X^t: M\to M, \quad \phi_X^t(q) = \gamma(t;0, q)$ where $t\in\bbR$.
In this case we say that the vector field generates a {\bf  flow}\footnote{Not to be confused with NF, a {\bf flow} is only defined for a subset $\mathcal{D}\subseteq \bbR\times M$ in general, since not all vector fields are complete.
A flow defined on all $\bbR\times M$ is often called a {\bf global flow}. For simplicity we restrict our attention to complete vector fields and global flows. In the rest of the paper a flow will denote a globally defined flow}.
\begin{defn}
A smooth  flow is a smooth left $\bbR$-action on a manifold $M$; that is, a family of smooth diffeomorphisms $\phi^t:M\to M, \forall t\in \bbR$ satisfying the following properties for all $s,t\in \bbR$ and $q\in M$: 
\begin{equation}
    \phi^0=\text{Id}, \quad  \phi^t \circ \phi^s(q) = \phi^{t+s} \ \forall t,s\in \bbR
\end{equation}
\end{defn}

Every smooth  flow $\phi$ uniquely generates a smooth complete vector field $X$ by $
    X_q = \frac{d}{dt}\bigr|_{t=0} \phi^t(q)\quad \forall q\in M $
Conversely every complete smooth vector field generates smooth  flow $\{\phi_X^t\}_{t\in\bbR}$ through its integral curves. This result is known as the {\bf Fundamental Theorem of Flows}. 

For time dependent vector field we have to take into account the additional time dependence, we therefore {\bf have time dependent  flows}
\begin{defn}
A time dependent smooth  flow on a smooth manifold is a two parameter family of diffeomorphism $\{\phi^{t_1,t_0}\}_{t_0,t_1\in \bbR}$, $\phi^{t_1,t_0}:M\to M, \forall t_0, t_1\in \bbR$
that satisfy the following conditions:
\begin{align}
  \phi^{t,t} = Id, \quad 
  \phi^{r,s}\circ \phi^{s,t} = \phi^{r,t} \quad \forall t, s, r \in \bbR
\end{align}
\end{defn} 
Similarly as the time independent case we have that a time dependent complete smooth vector field uniquely generates a time dependent smooth flows and vice versa. 

Summing up we have seen how vector fields naturally allow us define maps on a generic smooth manifold $M$. We refer to \cite{lee2013smooth} for proofs and additional results on vector fields and flows on manifolds.

\section{Cotangent lift}\label{sec:cot-lift}

Let $X\in\smoothsec{M}{TM}$ a complete smooth vector field \footnote{We can consider a time dependent vector field $Y:\bbR\times M \to TM$ as a vector field $\breve{Y}\in\smoothsec{\bbR\times M}{T\bbR\times TM}$ on the augmented space $\bbR\times M$: $\breve Y(s,q):= \lp \frac{\partial}{\partial t}|_{t=s}, Y(s, q)\rp$ } 
and $\{\phi^t_X\}_{t\in\bbR}$ its generated flow. We are interested in differentiating through $\phi^t_X: M\to M$. In differential geometry this corresponds to computing the pullback map $(\phi^t_X)^*: T^*M\to T^*M$, where $T^*M$ is the cotangent bundle of $M$. The key observations of the adjoint sensitivity method is that in $M = \bbR^n$ this quantity can be computed simulating the reverse dynamics of $X$, augmenting it with an additional linear ODE called the {\bf adjoint equation}. We will show how to generalize this method for vector fields on an arbitrary smooth manifold $M$. Consider the family of maps\footnotemark:
\begin{align}
    \lp\phi^t_{-X}\rp^*: T^*M &\to T^*M \qquad \forall t\in \bbR,\\
    p_q
    &\mapsto \lp\lp\phi^t_{-X}\rp^*p\rp_{\phi^t_{X}\lp q\rp}
\end{align}
Using\footnotetext{Given $p\in T^*M$ we use the notation $p_q$ to stress that $p$ has base point $q\in M$, this means $p\in T_q^*M$} the properties of the pullback and the fact that $\{\phi_{-X}^t\}_{t\in \bbR}$ is a flow it is easy to show that the family $\{\lp\phi^t_{-X}\rp^*\}_{t\in \bbR}$ defines a flow on $T^*M$. Following Section \ref{sec:vec-flow} there exists a unique vector field $\comp{X} \in \smoothsec{T^*M}{TT^*M}$ on the cotangent bundle that generates this flow, this vector field is called the {\bf cotangent lift} of $X$ \citep{GeoControlMech}:
\begin{align}
    \comp{X}_{p_q} = \frac{d}{dt}\biggr|_{t=0}\lp\lp\phi^t_{-X}\rp^*p\rp_{\phi^t_{X}\lp q\rp}\quad \forall p_q\in M
\end{align}
This means that given the cotangent vector $p_q\in T_q^*M$, to  compute the pullback of $p_q$ by $\phi_{X}^t$ we can solve the Cauchy problem defined by $-\comp{X}$ with starting point $p_q$.
The cotangent lift has the following important properties: \footnote{See Remark S1.11 in \cite{GeoControlMech}}
\begin{prope}\label{prope:lift1}
 $\comp{X}$ is a Hamiltonian vector field with respect to the canonical symplectic structure of the cotangent bundle $T^*M$. The Hamiltonian that generates $\comp{X}$ is $H_X(p_q) = p_q\lp X_q\rp$. We therefore have:
 \begin{align}
     \comp{X} \lrcorner\,  \omega = dH_X,
 \end{align}
 where $\omega$ is the canonical symplectic form on $T^*M$
\end{prope}
\begin{prope}\label{prope:lift2}
$\comp{X}$ is a linear vector field on the fibers of $T^*M$. That is, given $p_q, p'_q \in T^*_q M$, and $a,b\in \bbR$ it holds that $\comp{X}(a\!\cdot\! p_q + b\!\cdot\! p'_q) = a\!\cdot\!\comp{X}(p_q) + b\!\cdot\!\comp{X}(p'_q)$. \end{prope}
This fundamentally descends because the pullback map is fiberwise linear.
\begin{prope}\label{prope:lift3}
$\comp{X}$ is a lift of $X$. That is $d\pi_{M}(\comp{X}_{p_q}) = X_q, \forall p_q\in T^*M$. Where $\pi_M:T^*M\to M$ is the projection of the cotangent bundle into its base space M.
\end{prope}
\subsection{Cotangent lift in local coordinates}
Let's compute a local expression for the cotangent lift $\comp{X}$ on a local coordinate chart $(U; x_i), U\subseteq M$. In this chart the vector field will have expression $X|_U = \sum_{i=1}^nf_i\partial_{x_i}$ where $f_i \in C^{\infty}(U)$.
Since $\comp{X}$ is a vector field on $T^*M$ we can find its components with respect to the frame $\{\partial_{x_i}, \partial_{\xi_i}\}_{i=1}^n$ adapted to cotangent coordinates $(T^*U; x_i, \xi_i)$ \footnote{See Section 2.1 in \citep{da2001lectures}}. Since the field is Hamiltonian,  we can leverage the fact that an Hamiltonian vector fields $Y$ with Hamiltonian $H$ in local cotangent coordinates can be written using Hamilton equations:
\begin{align}
    Y\big|_{T^*U} = \sum_{i=1}\lp\dfrac{\partial H}{\partial \xi_i} \partial_{x_i} - \dfrac{\partial H}{\partial x_i} \partial_{\xi_i}\rp
\end{align}
In our specific case we have that the local expression for $H_X$ is:
\begin{align}
    H_X\big|_{T^*U} = \sum_{i=1}^n \xi_i dx_i\lp\sum_{j=1}^nf_j\partial_{x^j}\rp = 
    \sum_{i=1}^n \xi_i f_i
\end{align}
Therefore:
\begin{align}\label{eq:cotangent-lift-coord}
    \comp{X}\big|_{T^*U} = \sum_{i=1}^nf_i\partial_{x_i} -
    \sum_{i=1}^n\lp\sum_{j=1}^n \dfrac{\partial f_i}{\partial x_j}\xi_j\rp\partial_{\xi_i}
\end{align}
Notice that as we expected from Property \ref{prope:lift2} and \ref{prope:lift3}, cotangent lift \eqref{eq:cotangent-lift-coord} is linear on the components $\partial_{\xi_i}$ and coincides with $X$ if projected on the components $\partial_{x_i}$. Notice this expression is the same as the adjoint equation. Therefore for $M=\bbR^n$ the cotangent lift coincides with adjoint equation. 
\subsection{Cotangent lift on embedded submanifolds}
Let $M$ be a properly embedded smooth submanifold of $\bbR^m$, and let $\iota: M\hookrightarrow\bbR^m$ denote the inclusion map. Consider a smooth vector field $X\in \smoothsec{M}{TM}$ and $\overline{X}\in \smoothsec{\bbR^n}{T\bbR^n}$ a smooth tangent vector field that extends $X$. 
We first observe that since the $\overline{X}$ and $X$ are $\iota$-related their flows commute with the inclusion map (Proposition 9.6 \citep{lee2013smooth}). We therefore have that the following diagram commutes:
\begin{equation}\label{diag:embedding-commute}
\begin{tikzcd}[column sep=large]
 \bbR^m \arrow{r}{\phi^t_{\overline{X}}} \arrow[d, hookleftarrow,"\iota"]& \bbR^m\arrow[d, hookleftarrow,"\iota"] \\
 M \arrow{r}{\phi^t_{X}} & M
\end{tikzcd}
\end{equation}
Suppose we are interested in computing the differential of the function $f\circ \iota \circ \phi_X^t = f\circ \phi_X^t: M\to \bbR$ where $f:\bbR^m\to\bbR$. We can then both write:
\begin{align*}
d\lp f\circ \iota \circ \phi_X^t\rp &= \lp\phi_X^t\rp^* \circ \iota^*\circ df = \phi_{-\comp{X}}^t \circ \iota^*\circ df \\
&= \iota^*\circ \lp\phi_{\overline{X}}^t\rp^* \circ df = \iota^*\circ \phi_{-\comp{\overline{X}}}^t\circ df
\end{align*}
This means that we can use the cotangent lift of $\overline{X}$ to compute the pullback of cotangent vectors by the flow of $X$.

\section{Continuous normalizing flows on manifolds}\label{sec:cnf}
Let $(M, g)$ a orientable Riemannian manifold and $\mu_g\in \smoothsec{M}{\Lambda^nT^*M}$ its Riemannian volume form(\citet{lee2013smooth}, Proposition 15.29).
Additionally let $X$ a complete smooth time dependent vector fields on $M$ and $\{\phi^{t,s}_X\}_{t,s\in \bbR}$ its smooth time dependent flow. In Section \ref{sec:vec-flow} we saw that the maps $\phi^{t,s}_X:M\to M$ define a diffeomorphisms on $M$. We can then use the flow induced by $X$ to define continuous normalizing flows on $M$. 

We represent our initial probability density using the volume form $\rho_0\mu_g\in\smoothsec{M}{\Lambda^nT^*M}$
Where $\rho_0\in C^\infty(M)$ is a smooth non-negative function on $M$ that integrates to one $\int \rho_0\mu_g = 1$. 
To describe our reparameterized density we define $\rho_t\in C^\infty(M)$ as the smooth function such that:
\begin{equation}
    \rho_t\mu_g = \lp\phi^{0,t}_X\rp^*\lp\rho_0\mu_g\rp,
\end{equation}
where $(\phi^{0,t}_X)^*: \smoothsec{M}{\Lambda^nT^*M}\to \smoothsec{M}{\Lambda^nT^*M}$ is the pullback of volume forms induced by $\phi^{0,t}_X$(\citet{lee2013smooth}, Chapter 14). The evolution of $\rho_t$ over time is given by the {\bf continuity equation}( \citet{khalil2017types}, Section 4).
\begin{thm}[Continuity equation]\label{thm:continuity-equation}
Let $M$, $\mu_g$, $X$, $\rho_t$ as defined above. Then the function $\rho\in C^\infty(M\times\bbR),\ \rho(\cdot, t) := \rho_t$ satisfies the following linear PDE:
\begin{align}
    X_t(\rho_t) + \rho_t\dive{}{X_t} = - \partial_t \rho
\end{align}
\end{thm}
The divergence of a smooth vector field $Y\in \smoothsec{M}{TM}$ on a Riemannian manifold is the smooth function $\dive{}{Y}\in C^\infty(M)$ such that
\begin{align}
\dive{}{Y} \mu_g = \mathcal{L}_Y\lp\mu_g\rp = d\lp Y \lrcorner\, \mu_g \rp,
\end{align} 
where $\mathcal{L}_Y(\mu_g)$ is the Lie derivative of the the Riemannian volume form with respect to Y.
We are interested in computing how the value of $\rho_t$ changes on the flow curves $t\mapsto \phi_{X}^{t,0}(q_0), q_0\in M$. Using the continuity equation and the chain rule we have:
\begin{align*}
\frac{d}{dt}\left[\rho_t\lp\phi_{X}^{t,0}\lp q_0\rp\rp\right] = \bigg[- \dive{}{X_t}\rho_t\bigg]\lp\phi_{X}^{t,0}(q_0)\rp\ \ 
\end{align*}
If we fix a starting point $q_0\in M$ we obtain a linear ODE on $\bbR$. We can then solve for an initial value $\rho_0(q_0)$ of the probability density:
\begin{align*}
\rho_t\!\lp\phi_X^{t,0}(q_0)\rp =\exp\lp\! -\! \int_0^t \dive{}{X_t}\!\lp\phi_X^{t,0}(q_0)\rp dt \rp\! \cdot\! \rho_0(q_0)
\end{align*}
In many applications we are interested in the log probability density in which case the expression further simplifies to:
\begin{align*}
\log \rho_t\lp\phi_X^{t,0}(q_0)\rp =\log\rho_0(q_0) -\!  \int_0^t \dive{}{X_t}\!\lp\phi_X^{t,0}(q_0)\rp dt
\end{align*}
\section{Related Work}
As mentioned in the Introduction, the absence of a general procedure for parameterizing maps between manifolds has been the main obstacle in defining normalizing flows on manifolds. 
\citet{gemici2016normalizing} try to sidestep this by first mapping points from the manifold $M$ to $\bbR^n$, applying a normalizing flow in this space and then mapping back to $M$. However when the manifold $M$ has a nontrivial topology there exist no continuous and continuously invertible mapping, i.e. a \emph{homeomorphism} between $M$ to $\bbR^n$, such that this method is bound to introduce numerical instabilities in the computation and singularities in the density. Similarly \citet{falorsi2019reparameterizing} create a flexible class of distributions on Lie groups by mapping a complex density from the Lie algebra to the group using the exponential map. While the exponential map is not discontinuous, for some particular groups the resulting density can still present singularities when the initial density in the Lie algebra is not properly constrained. 
\citet{rezende2020normalizing} define normalizing flows for distributions on hyperspheres and Tori. This is done by first showing how to define diffeomorphisms from the circle to itself by imposing special constraints. The method is then generalized to products of circles, and extended to the hypersphere $S^n$, by mapping it to $S^1\times[-1,1]^n$ and imposing additional constraints to ensure that overall map is a well defined diffeomorphism. \citet{bose2020latent} define normalizing flows on hyperbolic space by successfully taking in to account the different geometry, however the definition of a diffeomorphisms in hyperbolic space is made easier due to the fact that {\it topologically} the hyperbolic space is homeomorphic to the Euclidean one.


\section{Conclusion and future work}
Future research will experiment with the presented framework in a wider range of tasks and manifolds. In addition we will explore how to further improve the scalability of the defined techniques. Possible directions are Monte Carlo approximations of the divergence on manifolds, using numerical integrators adapted to the specific manifold structure and regularization methods based on optimal transport on manifolds (in the spirit of \citet{finlay2020train}).

\clearpage
\newpage

\bibliography{bibliography.bib}
\bibliographystyle{icml2020}
\newpage
\onecolumn
\appendix 
\section{Proof of Concept Experiments}\label{app:experimments}
As a proof of concept, we show how the proposed Manifold Continuous Normalizing Flow (MCNF) is able to learn complex multi-modal densities on the hyper-sphere. As target densities we used the Mixture of von Mises-Fisher on $S^2$ and $S^3$ defined in \citet{rezende2020normalizing}.
Each model was optimized using Adam \cite{kingma2014adam} for 10000 epochs, learning rate of $10^{-3}$ and batch size of 256. See \citet{rezende2020normalizing} for a detailed description of the task and of the metrics used.

\begin{figure}[t]
\centering
\subfigure[Target]{\includegraphics[width=0.45\linewidth]{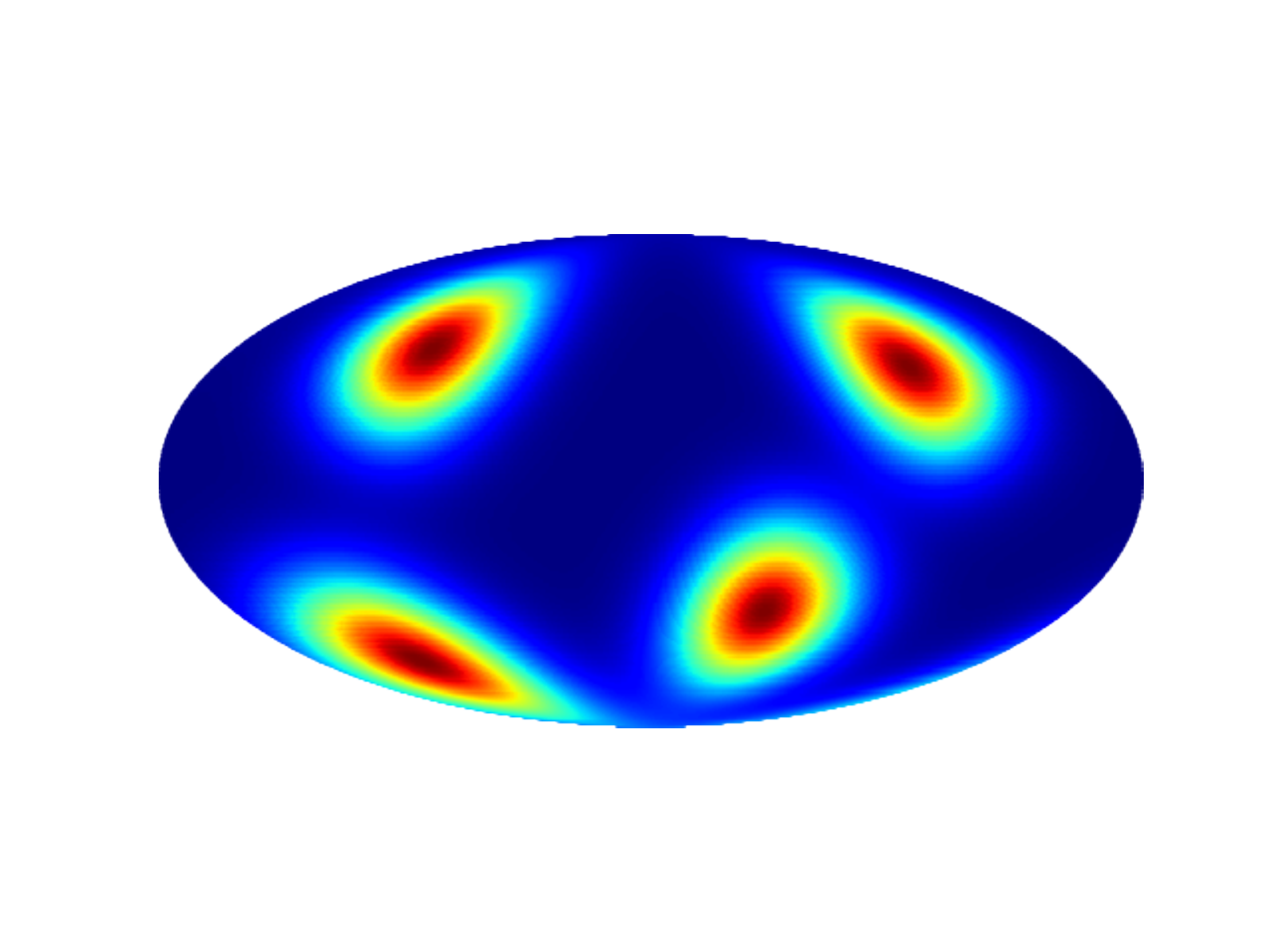}}
~
\subfigure[Model]{\includegraphics[width=0.45\linewidth]{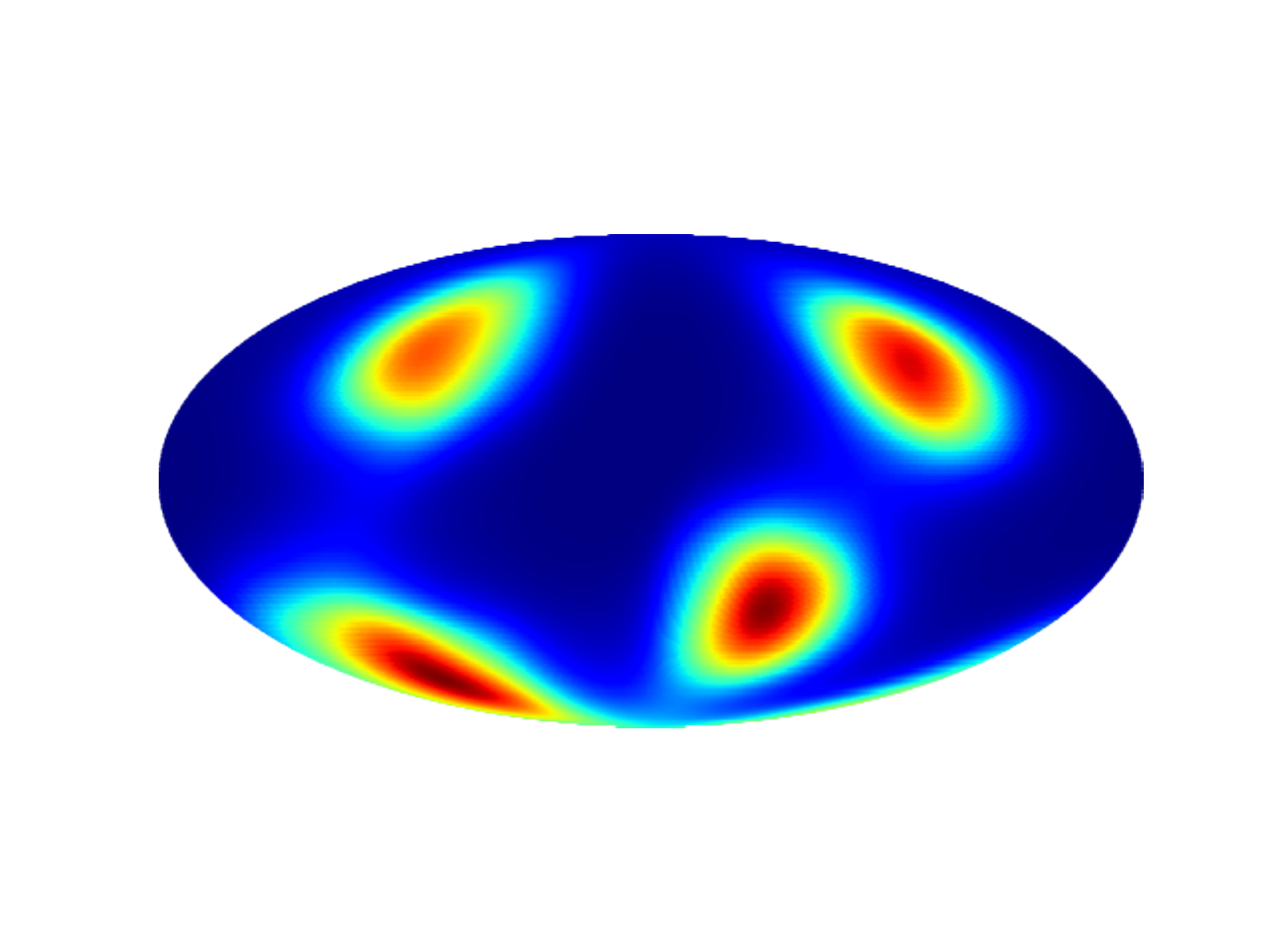}}
\caption{Learned density on $S^2$}
\label{fig:density-match-vmf}
\end{figure}

Results are reported in Table \ref{tab:match-sphere} while Figure \ref{fig:density-match-vmf} shows the leaned density on $S^2$. We observe that the proposed model is able to closely match the target densities, with a considerably lower KL divergence than the model by \citet{rezende2020normalizing}.

\begin{table}
\begin{center}
\begin{tabular}{c|l|c|c}
    \hline
    {\bf Manifold} & {\bf Model} & {\bf KL[nats]}& {\bf ESS[\%]} \\
    \hline \hline
    \multirow{2}{4em}{$S^2$} & MS($N_T = 1$, $K_m = 12$, $K_s = 32$)& .05{\tiny$\pm$.01} & 90\\
                        & MCNF($H=[10, 10]$) & \textbf{.008}{\tiny$\pm$.001} & \textbf{98.4}{\tiny$\pm$.2} \\
    \hline \hline
    \multirow{2}{4em}{$S^3$} & MS($N_T = 1$, $K_m = 32$, $K_s = 64$)& .14 & 84\\
                        & MCNF($H=[10, 10]$) & \textbf{.013}{\tiny$\pm$.001} & \textbf{97.5}{\tiny$\pm$.2} \\
                        
    \hline
\end{tabular}
    \caption{Evaluation of Manifold Continuous Normalizing Flow on Manifold (MCNF) on density matching task. Performance is measured using KL divergence end Effective Sampling Size. $H$ indicates the hidden units in each hidden layer. Results are compared with recursive M\"obius-spline flow (MS) \citep{rezende2020normalizing}. $N_T$ is the number of stacked transformations for each flow; $K_m$ is the number of centres used in Mobius;  $K_s$ is the number of segments in the spline flow. Error is computed over 3 replicas of each experiment}
    \label{tab:match-sphere}
\end{center}
\end{table}

\section{Parameterizing vector fields on manifolds}\label{app:par-field}
Given a manifold $M$ we are left with the problem of parameterizing a large enough set of vector fields that allows to express a rich class of distributions on the manifold.  When we try to parameterize a large set of function we look at neural networks as a natural solution, however they can only parameterize functions $\bbR^n\to\bbR^m$, and therefore there is no straightforward way to use them.
Finding the best way of parameterizing vector fields on manifolds is an interesting problem with no unique solution, how to tackle it will largely depend on how the manifold is defined and what data structure is used to parameterize it in practice. Nevertheless all the objects and methods discussed in the rest of the paper are defined independently from the specific parameterization method chosen. Therefore, if in the future a better way of parameterizing vector fields will emerge, they will still be applicable.

Notwithstanding the above, in this section we will try to give some guidance on how to approach this problem. In the first part we will show how, using generators, it can be reduced to the much easier task of parameterizing functions on manifolds. We will then give some practical advice in the case where the manifold is described using an embedding in $\bbR^m$. Throughout this section, given a function $f:M \to \bbR^m$ we will indicate with $f_i:M\to \bbR$ its $i$-th component, such that $f = (f_1, \cdots, f_m)$.

\subsection{Local frames and global constraints}\label{sec:vector-fields-generators}
We begin by analyzing how we parameterize vector fields in $\bbR^n$, to investigate to what extent we can generalize this procedure. In the euclidean space vector fields are simply functions $f:\bbR^n\to\bbR^n$. In a more geometrical language the function $f$ defines the vector field $X$ in the following way:
\begin{align}\label{eq:field-Rn}
  X = f_1\partial{x^1} + \cdots +  f_n\partial{x^n} 
\end{align}
The converse is also true: \emph{for every vector field $X$ there exist a unique continuous function $f:\bbR^n\to\bbR^n$ such that Equation \eqref{eq:field-Rn} holds}. On a generic $n$-dimensional smooth manifold this is only true {\it locally}. This means that there exists a open cover $\{U_i\}_{i\in \mathcal{I}}$ of $M$\footnote{Assuming that the manifold is second countable, there exists  $\mathcal{I}$ that is finite and has cardinality $n+1$, see Lemma 7.1 in \cite{metric-structures-diff}}, called the {\bf trivialization cover}, such that $TM$ restricted to each $U_i$ is isomorphic to the trivial bundle. This is equivalent to saying that for every set $U_i$ there exist $n$ smooth vector fields $E_1^{(i)}, \cdots, E_n^{(i)}\in \smoothsec{U_i}{TU_i})$ such that for every smooth vector field $X\in \smoothsec{M}{TM}$ there exists a unique smooth function $f: U_i\to \bbR^n$ such that:
\begin{align}
    X \bigr|_{U_i}= f_1E_1^{(i)}+ \cdots+ f_nE_n^{(i)}
\end{align}
We then can call $E_1^{(i)}, \cdots, E_n^{(i)}$ a {\bf local frame}. A local frame that is defined on an open domain $U=M$ (this means on the entire manifold) is called a {\bf global frame}. On a manifold there exists plenty of local frames, in fact given a smooth local chart $(U, \{x^i\})$ the fields $\partial_{x^1},\cdots, \partial_{x_n} \in \smoothsec{U}{TU}$ form a local frame called {\bf coordinate frame}. In the special case of $\bbR^n$ its coordinate frame is a global frame. Unfortunately in general not every manifold has a global frame, the simplest example is the sphere $S^2$. In the sphere case it is well known that there exists no vector field that is everywhere nonzero, this result goes by the {\it hairy ball theorem}. 
It is then clear that no pair of vector fields $E_1, E_2 \in \smoothsec{M}{TM}$ can form a global frame, in fact there will always be a point $q\in S^2$ such that: 
\begin{align}
    \text{span}\lp \lp E_1\rp_q, \lp E_2\rp_q \rp \le 1 < 2 = \text{dim}\lp T_q M\rp
\end{align}
The manifolds for which a global frame $E_1, \cdots, E_n\in \smoothsec{M}{TM}$ exists are called {\bf parallelizable manifolds}, for this class we can parameterize all smooth vector fields on $M$ in the same way as we did on $\bbR^n$. This means choosing a smooth function $f:M\to \bbR$ and defining a vector field $X$:
\begin{align}
    X = f_1E_1+ \cdots+ f_nE_n
\end{align}
A manifold is parallelizable iff its tangent bundle is isomorphic to the trivial bundle: $\bbR^n\times M\simeq TM$. A global frame gives an explicit isomorphism:
\begin{align*}
\bbR^n\times M&\to TM\\
(z,q)&\mapsto z_1\lp E_1\rp_q +\cdots +  z_n\lp E_n\rp_q
\end{align*}
An important and large class class of parallelizable manifolds is given by Lie Groups, which are smooth manifold which additionally posses a group structure compatible with the manifold structure. For background on Lie Groups we refer to \cite{lee2013smooth}
\subsection{Generators of vector fields}
We have seen that for parallelizable manifolds, once we have defined a global frame, we have a bijective correspondence between functions $C^\infty(M, \bbR^n)$ and smooth vector fields:
\begin{align}
    C^\infty(M, \bbR^n) &\to \smoothsec{M}{TM}\\
    f&\mapsto f_1E_1+ \cdots+ f_nE_n
\end{align}
For non parallelizable manifolds, we fail to find a global frame because given any $n$ vector fields $\{E_i\}_{i=1}^n$
there always exist points $q$ where all $\{\lp E_i\rp_q\}_{i=1}^n$ fail to span all $T_qM$:
\begin{align*}\label{eq:func-field-frame}
    \text{span}\lp\{\lp E_i\rp_q\}_{i=1}^n\rp \subsetneq T_qM
\end{align*}
The idea is then to add vector fields to the set $\{\lp E_i\rp_q\}_{i=1}^n$, giving up on the injectivity, until they "generate" all $\smoothsec{M}{TM}$. To make this statement more precise we have to use the language of {\bf modules}. In fact in general the space of smooth sections of a vector bundle $(E, \pi, M)$ forms a module over the ring $C^\infty(M)$ of the smooth functions on $M$.
\begin{defn}
A finite set of vector fields $\{X_i\}_{i=1}^m\subset \smoothsec{M}{TM}, m\in \bbN_{>0}$ is a generator of the $C^\infty(M)$-module of the smooth vector fields on $M$ if for every vector field $X\in \smoothsec{M}{TM}$ there exist 
$\{f_i\}_{i=1}^m\subset C^\infty(M)$ such that:
\begin{align}
    X = f_1X_1+ \cdots+ f_mX_m
\end{align}
If there exist a generator for for $\smoothsec{M}{TM}$ we then say that $\smoothsec{M}{TM}$ is finitely generated.
\end{defn}
\begin{thm}\label{thm:exists-generator}
Let M be a (second countable) smooth manifold $M$. Then the module of smooth vector fields $\smoothsec{M}{TM}$ is finitely generated.
\begin{proof}
Since $M$ is second countable we can apply Lemma 7.1 in \cite{metric-structures-diff} and say that there exist an open trivialization cover $\{U_i\}_{i=0}^n$, where $n$ is the dimension of $M$. We denote with $E_1^{(i)},\cdots, E_n^{(i)}$ the local frame relative to the domain $U_i\subseteq M$. Now let $\{\psi_i\}_{i=0}^n$ be a smooth partition of unity subordinate to $\{U_i\}_{i=0}^n$. We define the global vector fields on $M$:
\begin{align}
\grave{E}^{(i)}_{j} := \begin{cases} \psi_i\cdot E^{(i)}_{j}, & \mbox{on  } U_i \\ 0 & \mbox{on  } M\setminus\mbox{supp}\psi_i \end{cases}
\quad \forall i\in\{0,\cdots,n\}\ \forall j\in\{1,\cdots,n\}
\end{align}
We  have that $\{\grave{E}^{(i)}_{j}\}_{\substack{i=0..n\\j=1..n}}$ is a generator of $\smoothsec{M}{TM}$.
To prove this we first define $V_i\subseteq M$ as the open set where $\psi_i>0$. Since $\sum_{i=1}^n\psi_i = 1$ then also $\{V_i\}_{i=0}^n$ is an open cover of $M$. Given a global global smooth vector field $X\in \smoothsec{M}{TM}$, for each $V_i$ there exist $f^{(i)}_1, \cdots, f^{(i)}_n\in C^\infty(V_i)$ such that:
\begin{align}\label{eq:local-frame-equality}
    X =_{V_i} \sum_{j=1}^n f^{(i)}_j\grave{E}^{(i)}_{j} \quad \forall i\in\{1,\cdots,n\}
\end{align}
Now let $\{h_i\}_{i=0}^n$ a smooth partition of unity subordinated to $\{V_i\}_{i=0}^n$. We use it to define the global smooth functions:
\begin{align}
\grave{f}^{(i)}_{j} := \begin{cases} h_i\cdot f^{(i)}_{j}, & \mbox{on  } V_i \\ 0 & \mbox{on  } M\setminus\mbox{supp}h_i \end{cases}
\quad \forall i\in\{0,\cdots,n\}\ \forall j\in\{1,\cdots,n\}
\end{align}

Combining equation \eqref{eq:local-frame-equality} and with the fact that $\sum_{i=1}^n h_i = 1$ we have:
\begin{align}
    X = \sum_{i=0}^n\sum_{j=1}^n \grave{f}^{(i)}_{j} \grave{E}^{(i)}_{j} 
\end{align}
\end{proof}
\end{thm}
From this Theorem and the definition of generator we can extract a methodology to parameterize all vector fields on smooth manifolds:
\begin{enumerate}
    \item choose a suitable set of generators $\{ X_i\}_{i=1}^m$
    \item find a way of parameterizing functions $f_i:M\to \bbR$
    \item model a generic vector field $X$ as a linear combination:
    \begin{align}\label{eq:vec-field-gen}
        X = f_1X_1+ \cdots+ f_mX_m
    \end{align}
\end{enumerate}
The above proof also tells us that a simple and general recipe to obtain a generator is to take a collection of local frames and multiply them by a smooth partition of unity. 
The efficiency of this framework is given by the cardinality of the generator: lower cardinality requires parameterization of less functions. The proof gives us an initial upper bound on the lowest cardinality of the set of generators we can achieve for a generic manifold: $n^2+n$ where $n$ is the dimension of the manifold. We will see that for Riemannian manifolds, using the Whitney embedding theorem this number can be further reduced to $2n + 1$. 
\newline 
However the cardinality of the generator is not the only factor to consider when choosing a good generating set. In fact combining equation \eqref{eq:vec-field-gen} with the properties of Lie derivative we obtain:
\begin{align}
    \dive{}{X} = \sum_{i=1}^m X_i(f_i) + f_i\dive{}{X_i}
\end{align}
If we can find a set of generators with known divergence, or for which we can (pre-)compute the divergence, this greatly simplifies the divergence computation. 
\newline
\subsubsection{Time dependent vector fields}
When parameterizing time dependent vector fields vector fields we have to model a vector field $X_t$ for all $t\in\bbR$. Using generators we can easily accomplish this by parametrizing a function $f:\bbR\times M\to\bbR^m$ and defining
\begin{align}
    X_t:= f_1(t,\cdot)X_1 + \cdots + f_m(t,\cdot)X_m
\end{align}

\subsection{Embedded submanifolds of $\bbR^m$}
\begin{defn}
Let $N$, $M$ smooth manifolds and $F:M\to N$ a smooth function between them. Given $X\in \smoothsec{M}{TM}$ and $Y\in \smoothsec{N}{TN}$ smooth vector fields respectively on $M$ and $N$, we say that they are {\bf F-related} if
\begin{align}
    dF_p\lp X_p\rp = Y_{F(p)} \quad \forall p\in M
\end{align}
\end{defn}
A general way to work in practice with manifolds is using embedded submanifolds of $\bbR^m$. An embedding for a manifold $M$ is a continuous injective function $\iota: M\hookrightarrow \bbR^m$ such that $\iota: M\to \iota(M)$ is a homeomorphism. The embedding is smooth if $\iota$ is smooth and $M$ is diffeomorphic to its image. In this case $\iota(M)$ is a smooth submanifold of $\bbR^m$. For all practical purposes we can directly identify $M$ as a submanifold of $\bbR^m$, the function $\iota: M\hookrightarrow \bbR^m$ then simply denotes the inclusion. Through this identification we can then consider the tangent space $T_qM,\ q\in M$ as a vector subspace of $T_q\bbR^m$. An embedding is said {\bf proper} if $\iota(M)$ is a closed set in $\bbR^m$.\footnote{Requiring that the embedding is proper excludes embeddings of the form $U\hookrightarrow M$ where $U$ is an open subset of $M$}
\begin{thm}[Whitney Embedding Theorem, 6.15 in \cite{lee2013smooth}]
Every smooth $n$-dimensional manifold admits a proper smooth embedding in $\bbR^{2n+1}$
\end{thm}
The Whitney embedding theorem tells us that parameterizing manifolds as submanifolds of the Euclidean space gives us a general methodology to work with manifolds. Developing algorithms that assume that the manifold is given as an embedded submanifold of $\bbR^m$ is therefore of outstanding importance. 

For embedded submanifolds parameterizing functions is extremely easy, and can be simply done via restriction: given a smooth function $f:\bbR^m\to \bbR$, $f\circ\iota$ then defines a smooth function from $M$ to $\bbR$. 

Unfortunately for vector fields it is not as easy as for functions, in fact in general given a vector field $X\in C(\bbR^m,T\bbR^m)$ this does not restrict in general to a vector field on a submanifold $M\subseteq \bbR^m$, as in general given $q\in M$ we have $X_q\not\in T_qM\subseteq T\bbR_q^m$. In order for $X$ to restrict to a vector field on a submanifold $M$ we need for $X$ to be {\bf tangent to the submanifold} :
\begin{align}
    X_q\in T_qM\subseteq T_q\bbR^m\quad \forall q\in M
\end{align}
A tangent vector field then defines a vector field on the submanifold:

\begin{lem}
Let $M$ be a smoothly embedded submanifold of $\bbR^m$, and let $\iota: M\hookrightarrow\bbR^m$ denote the inclusion map. If a smooth vector field $Y\in \smoothsec{\bbR^m}{T\bbR^m}$ is tangent to $M$
there is a unique smooth vector field on $M$, denoted by $Y|_M$ , that is $\iota$-related to $Y$. Conversely a vector field $\overline{Y}\in \smoothsec{\bbR^m}{T\bbR^m}$ that is $\iota$-related to $Y$ is tangent to $M$ 
\begin{proof}
See proof of Proposition 8.23 in \cite{lee2013smooth}
\end{proof}
\end{lem}
More importantly we can parameterize all vector fields on an embedded submanifold using tangent vector fields:
\begin{prop}
Let $M$ be a properly embedded submanifold of $\bbR^m$, and let $\iota: M\hookrightarrow\bbR^m$ denote the inclusion map. For any smooth vector field $X\in \smoothsec{M}{TM}$ there exist a smooth vector field $\overline{X}\in \smoothsec{\bbR^m}{T\bbR^m}$ tangent to $M$ such that:
\begin{align}
     \overline{X}|_M = X
\end{align}
We call $\overline{X}$ an extension of $X$.
\begin{proof}
Let $U\subseteq\bbR^m$ be a tubular neighborhood of $M$, then by Proposition 6.25 of \cite{lee2013smooth} there exist a smooth map $r:U\to M$ that is both a retraction and a smooth submersion. Then since $r$ is a submersion there exist a vector field $\overline{X}\in \smoothsec{U}{TU}$ that is $r$-related to $X$ \footnote{See for example Exercise 8-18 of \cite{lee2013smooth}}. This means $ dr_z \overline{X}_z = X_{r(z)} \forall z\in \bbR^m$. Since $r$ is a retraction \begin{align}
d\iota_q\circ dr_q = Id_{T_q\bbR^m}\   
\Rightarrow\ \lp d\iota_q\circ dr_q\rp\overline{X}_q = \overline{X}_q
\Rightarrow\ d\iota_q X_q = \overline{X}_q\ \forall q\in M,
\end{align}
$\overline{X}$ is $\iota$-related to $X$ and therefore tangent to $M$ and such that $\overline{X}|_M = X$. Then $\overline{X}$ can be used to define a tangent vector field on all $\bbR^m$ using a smooth partition of unity subordinate to the open cover $\{\bbR^m\setminus M, U\}$.
\end{proof}
\end{prop}
From the proof of the theorem it's clear that the extension of $X$ is not unique. Our objective is then finding a way to parameterize all vector fields tangent to a submanifold. We first observe that given smooth vector fields $X,Y\in \smoothsec{\bbR^m}{T\bbR^m}$ tangent to $M$ and smooth functions $f,g\in C^\infty(\bbR^m)$ then $fX+gY$ is tangent to $M$. This means that the set of all smooth vector fields tangent to $M$ is a {\bf submodule} of the module of smooth vector fields on $\bbR^m$. Following the framework outlined in Section \ref{sec:vector-fields-generators} we then need to find $l$ tangent vector fields $\overline{X}_1,\cdots,\overline{X}_l$ such that $\overline{X}_1|_M,\cdots,\overline{X}_l|_M$ generates all $\smoothsec{M}{TM}$\footnote{Since the extension of a vector field is not unique this is different from finding a set of generators for the submodule of vector fields tangent to $M$}.
\subsubsection{Embedded Riemannian Submanifolds}
If our embedded submanifold manifold is equipped with a Riemannian metric, the gradient of the embedding gives us a set of generators for the tangent bundle. We first prove the following lemma
\begin{lem}\label{lem:pullback-emb-surj}
Let $M$ be a embedded submanifold of $N$, and let $\iota: M\hookrightarrow N$ denote the inclusion map. Then 
\begin{align}
    \iota^*: \iota^*T^*N &\to T^*M \\
    \beta_{\iota(q)}&\mapsto \iota^*\beta_{\iota(q)}:\quad v\mapsto \beta_{\iota(q)}(dv_q)\quad \forall q\in M,\ \forall \beta \in T^*_{\iota(q)} N,\ \forall v_q\in T_qM 
\end{align}
is a surjective vector bundle homomorphism, where by $\iota^*T^*N$ we denote the pullback bundle $\iota^*T^*N = \{(q,\beta)\in M\times T^*N| \pi(\beta)=\iota(q)\}$
\begin{proof}
Fix $q\in M$.Let $n$ be the dimensionality of $M$ and $m$ the dimensionality of $N$. We need to prove that $\iota^*:T^*_{\iota(q)}N\to T^*_qM$ is surjective.  Let $e_1,\cdots,e_n$ a basis for $T_qM$ and $\eta_1,\cdots, \eta_n$ its dual basis. By the linearity of $\iota^*$ it's then sufficient to prove that there exists $\beta_1,\cdots,\beta_n\in T^*_{\iota(q)}N$ such that for all $i\in\{1,\cdots,n\}$:
\begin{align}\label{eq:pullback-surjective-basis}
    \iota^*\beta_i = \eta_i
\end{align}
To see this, consider the set $\{d(\eta_1)_q,\cdots,d(\eta_n)_q\} \subset T_{\iota(q)}N$. Since $\iota$ is an embedding, $d\iota$ is injective. Therefore the vectors are linearly independent. 
We can then complete them to a basis $v_1:=d(\eta_1)_q,\cdots,w_n:=d(\eta_n)_q, w_{n+1},\cdots w_m$ of $T_{\iota(q)}N$. Let $\beta_1,\cdots,\beta_m \in T^*_{\iota(q)}N$ the dual basis. We then have:
\begin{align}
    \iota^*\beta_i(e_j) = \beta_i\lp d\lp\eta_j\rp_q\rp = \beta_i\lp w_j\rp = \delta_{ij} = \eta_i\lp e_j\rp \quad \forall i,j\in\{1,\cdots,n\}
\end{align}
Thus $\beta_i$ satisfies Equation \eqref{eq:pullback-surjective-basis} $\forall i\in\{1,\cdots,n\}$
\end{proof}
\end{lem}
\begin{thm}\label{thm:gen-gradient}
Let $(M,g)$ be a embedded submanifold of $\bbR^m$, and let $z: M\hookrightarrow \bbR^m$ denote the inclusion map. Then $\{\nabla z_i\}_{i=1}^m$ is a set of generators for smooth vector fields $\smoothsec{M}{TM}$. Where $\nabla$ denotes the Riemannian gradient with respect to the metric $g$. 
\begin{proof}
 Consider the differential forms $\{dz_i\}_{i=1}^m\subset \smoothsec{M}{T^*M}$. Using Lemma \ref{lem:pullback-emb-surj} we have that $\text{span}\lp\{dz_i(q)\}_{i=1}^m\rp = T^*_qM$, which means that at every point they span the cotangent space at the point. Using the musical isomorphism, this implies that the riemannian gradients $\nabla z_i$ span the tangent space at every point: 
 $\text{span}\lp\{\nabla z_i(q)\}_{i=1}^m\rp = T_qM$. 
 From this we can conclude that $\{\nabla z_i\}$ is a generator for $\smoothsec{M}{TM}$. 
 To see this consider the open sets $U_I:=\{q\in M | \{\nabla z_i(q)\}_{i\in I}\quad \text{are linearly independent}\}$ where $I\subset \{1,\cdots, m\}$ is any subset of indices of cardinality $n$. To see that these sets are open, observe that in a local coordinate chart $(U, (x_i))$ we can write $\nabla z_{i_k}|_U = \sum_{j=1}^n a^I_{i,j}\partial_{x_j}  \forall i\in {1,\cdots,M}$. In local coordinates the linear independence of $\{\nabla z_i\}_{i\in I}$, is equivalent to $\text{det}(A(x))\neq 0$. Where if $I={i_1,\cdots,i_n}$ $A(x)$ is defined as $A(x)_{jk} = a^I_{i_k,j}$. From the definition of $U_I$ it descends that the family $\{(U_I | I\subset \{1,\cdots,m\},\ \#(I)=n,\ U_I  \text{is not empty}\}$ forms a open trivialization. We can then conclude proceeding as in the proof of Theorem \ref{thm:exists-generator}
\end{proof}
\end{thm}
In general, given a function $f\in C^\infty(M)$ on a Riemannian manifold, its Laplacian is defined as the divergence of its Riemannian gradient:
\begin{align}
    \Delta f := \dive{}{\nabla f}
\end{align}
Then the divergence of the fields defined in Theorem \ref{thm:exists-generator} is given by the Laplacian of the functions $z_i:M\to\bbR$.

\subsubsection{Isometrically embedded Submanifolds}
If the manifold $M$ is isometrically embedded in $\bbR^n$.
Then $\nabla z_i$ is simply given by the orthogonal projection of the constant coordinate field $e_i = \partial_{x_i}$ from $T\bbR^m$ to $TM$. 
In this case the Laplacian of the functions $z_i:M\to \bbR$ is given by the mean curvature (\citet{chen2013laplace}, Proposition 2.3):
\begin{align}
\Delta z = nH = \text{tr}\ \sff,
\end{align}
where $H$ is the mean curvature and $\sff$ is the second fundamental form.

For the hypersphere $S^n$, the projection expression is particularly simple. This gives us $n+1$ vector fields $\{\overline{\nabla z_i}\}_{i=1}^{n+1}\subset \smoothsec{\bbR^{n+1}}{T\bbR^{n+1}}$ tangent to $S^n$ such that their restriction to $M$ forms a generator for $\smoothsec{S^{n}}{TS^n}$.  
\begin{align}
&\overline{\nabla z_i} = e_i - \langle z, e_i\rangle z \quad \forall i\in \{1,
\cdots, n+1\}\\
&\Delta z = -nz
\end{align}

\end{document}